# Process-Aware Procurement Lead Time Prediction for Shipyard Delay Mitigation

Yongjae Lee[1,†] [0009-0000-3418-6681], Eunhee Park[1,†], Daesan Park[1],
Dongho Kim[2], Jongho Choi[2] and Hyerim Bae[1,*]

[1] Pusan National University, Busan, Republic of Korea
[2] Samsung Heavy Industry Co., Ltd., Geoje, Republic of Korea
{yongzzai1102, eunheepark, ajtks9414, hrbae}@pusan.ac.kr
{d-ho.kim, j-ho.choi}@samsung.com

**Abstract.** Accurately predicting procurement lead time (PLT) remains a challenge in engineered-to-order industries such as shipbuilding and plant construction, where delays in a single key component can disrupt project timelines. In shipyards, pipe spools are critical components; installed deep within hull blocks soon after steel erection, any delay in their procurement can halt all downstream tasks. Recognizing their importance, existing studies predict PLT using the static physical attributes of pipe spools. However, procurement is inherently a dynamic, multi-stakeholder business process involving a continuous sequence of internal and external events at the shipyard, factors often overlooked in traditional approaches. To address this issue, this paper proposes a novel framework that combines event logs, dataset records of the procurement events, with static attributes to predict PLT. The temporal attributes of each event are extracted to reflect the continuity and temporal context of the process. Subsequently, a deep sequential neural network combined with a multi-layered perceptron is employed to integrate these static and dynamic features, enabling the model to capture both structural and contextual information in procurement. Comparative experiments are conducted using real-world pipe spool procurement data from a globally renowned South Korean shipbuilding corporation. Three tasks are evaluated, which are production, post-processing, and procurement lead time prediction. The results show a 22.6% to 50.4% improvement in prediction performance in terms of mean absolute error over the best-performing existing approaches across the three tasks. These findings indicate the value of considering procurement process information for more accurate PLT prediction.

## 1 Introduction

Rapid operational response to procurement delays is crucial in engineered-to-order industries such as shipbuilding and plant construction [1]. These projects synchronize numerous interdependent components, and a single bottleneck can lead to costly schedule disruptions that affect space, labor, and equipment utilization [2]. This vul-nerability is more pronounced in shipyards, where the delayed arrival of key components like pipe spools can stall entire assemblies and significantly increase costs. Pipe spools with flanges and valves form internal circulation systems and link critical systems, including propulsion, fuel, and fire-fighting systems [3]. In plant construction, pipe installation represents 43% of the total construction work and is considered a key stage on the critical path from a process management perspective [4]. In shipbuilding, these spools are installed deep within hull blocks soon after steel erection. Therefore, any delay in their delivery can significantly increase idle labor hours and rework in downstream tasks [5]. Consequently, accurate prediction of pipe spool procurement lead time (PLT) is a strategic priority for shipyards.

Despite extensive research in production planning, PLT prediction in practice often relies on heuristics. Recent studies [5, 6] have sought to automate PLT estimation using machine learning models trained on static attributes of spools (e.g., diameter, weight). However, pipe spool procurement is a complex, interconnected process involving multiple stakeholders and sequential events [7]. Although PLT is recorded after formal procurement shifts outside the shipyard, valuable process-related data begin accumulating internally much earlier. Therefore, focusing solely on static attributes reduces the analysis to fragmented information.

With the recent development of process-aware information systems, every time-stamped event in the workflow, even in shipyards, is recorded in a hierarchically structured dataset called the event log [8]. The recorded events include business process milestones (e.g., work order placement) and the corresponding attributes (e.g., department in charge). Existing approaches share two key limitations: 1) they ignore the information that could be extracted from the business process, and 2) they overlook temporal information, such as elapsed time.

This paper proposes a process-aware framework for predicting the pipe spool PLT to address these issues. The framework draws on the static spool attributes used in earlier studies and features extracted from the procurement event log (i.e., dynamic context). This dynamic context includes 1) the sequence of events, which captures the sequential nature of the process, and 2) time-related features for each event. As static data and dynamic context behave differently, they are processed by a non-sequential and sequential model, respectively, and then combined to make the final prediction. The effectiveness of the proposed method is validated by comprehensive experiments conducted on data from a globally renowned South Korean shipbuilding corporation. The result shows that our process-aware framework significantly improves predictive performance across production, post-processing, and procurement lead time predictions, underscoring the importance of process-aware information in shipyards.

The remainder of this paper is organized as follows. Section 2 provides background information, Section 3 introduces our proposed method, Section 4 shows experimental results and Section 5 conclude the research.

## 2  Background

### 2.1  Procurement Process

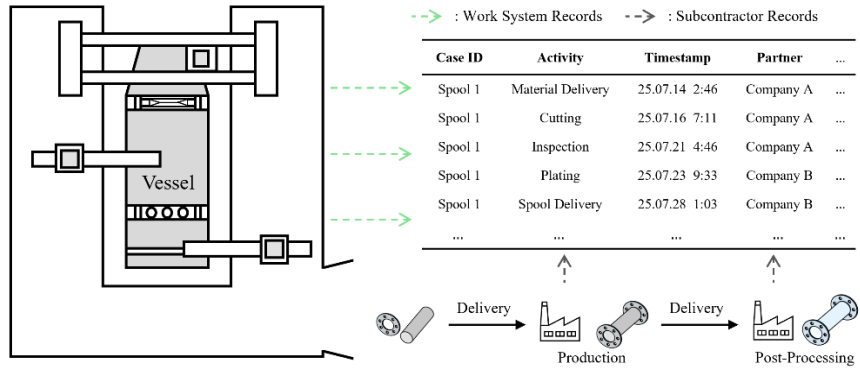

**Fig. 1.** The illustration of pipe procurement process and event log collection during the process.

Fig 1. illustrates the procurement and subcontracting process for pipe spools into four generalized steps. Once materials are secured, work orders are issued by spool unit, followed by a fabrication sequence consisting of material preparation, cutting, fit-up, welding, and inspection. Subsequently, finished products are returned in-house for storage and released to the installation stage as required. Notably, the handover sections between in-house and subcontractors (transportation, waiting, and inspection) constitute a critical area where variability accumulates, and procurement control in these transitions directly affects overall performance [5]. During this procurement process, the activities performed and the corresponding information (e.g., machine number, work location, etc.) are recorded in the log data together with timestamps.

In this study, the procurement lead time is defined as the elapsed time between the completion of raw material release and the receipt of the finished product. Accordingly, PLT encompasses not only the actual subcontracted processing time but also delays caused by transportation, waiting, re-inspection, and rework occurring during the transfer between in-house operations and subcontractors.

### 2.2  Preliminaries

In this section, the basic concepts, definitions, and notations for better understanding of our proposed framework are formalized.

**Event Log.** An event log [8] is a hierarchical-structured data that records the execution of business processes. In an event log, an *event* is a unit record of an occurrence that has the attributes of a *case*, an *activity*, *timestamp* and corresponding *attributes*. Partially borrowed from [9], event, trace, and log can be defined as:

**Definition 1 (Event, Trace, Log).** Let $\mathbb{E}$ be the event universe. An event is a tuple $e = (c, a, t, d_1, \ldots, d_f) \in \mathbb{E}$, denoting a unit record of process execution, where $c$ is the case identifier, $a$ is the activity, $t$ is the timestamp, and $d_f$ is the $f$-th corresponding attribute. A trace $\sigma = \langle e_1, e_2, \ldots, e_n \rangle \in \mathbb{E}^*$ is a non-empty finite sequence of events such that each event $e$ occurs only once. $\pi$ is a function for mapping attributes, such that $\pi_c(e) = c$. An event log $L = \{\sigma_1, \sigma_2, \ldots, \sigma_{|L|}\}$ is a set of traces, such that each event appears at most once in the log.

**Sequential Models.** To reflect the sequential nature of event log, we utilize sequential deep learning architecture family, which is basically designed to process the sequential structured data. For clarity, in this section, we provide explanations based on one of the most representative sequential models named Long-Short Term Memory (LSTM) [10]. Let $X_S = \{x_1, x_2, \ldots, x_t\}$ be the input sequence data where $T = \{1, 2, \ldots, t\}$ is the index of time steps are. By passing through $X_S$ to LSTM, each data at $t$-th step can be converted into meaningful representations by:

$$i_t, f_t, o_t, \tilde{c}_t = \sigma(W \cdot x_t + U \cdot h_{t-1} + b) \quad (1)$$

$$c_t = f_t \odot c_{t-1} + i_t \odot \tilde{c}_t \quad (2)$$

$$h_t = o_t \odot \tanh(c_t) \quad (3)$$

$$c_t, h_t = LSTM(x_t) \quad (4)$$

where $i_t, f_t, o_t$ are, respectively, the input, forget, and output gates, each constrained by [0,1] by the sigmoid $\sigma(\cdot)$. $\tilde{c}_t$ is the cell state candidate and $c_t$ is the updated cell state. $\odot$ represents the element-wise product. Thus, the input at $t$-th step $x_t$ is converted into cell state $c_t$ and hidden representation $h_t$. In the case of a bidirectional LSTM (Bi-LSTM) [10], the same input sequence is processed in two directions: a forward LSTM that propagates information from past to future, and a backward LSTM that propagates information from future to past.

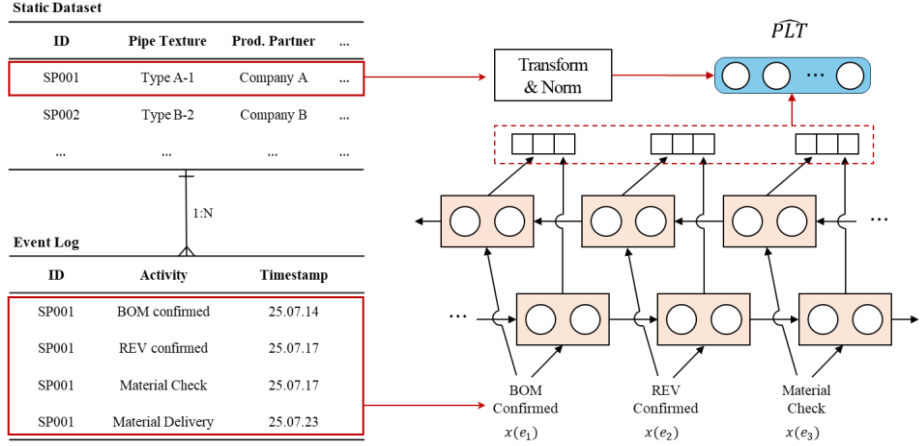

**Fig. 2.** The illustration of bidirectional sequential architecture for predicting PLT using collected static dataset and event log.

## 3 Proposed Method

Fig. 2. illustrates the generalized approach for predicting PLT based on static datasets and event logs. First, the feature of each pipe spool is transformed and normalized. Subsequently, each event of a trace (red box at bottom left) is sequentially processed. To this end, the features extracted from a single event, denoted by $x(e_i)$, are used as an input at each time step of the sequential model.

### 3.1 Feature Engineering

This study employs two types of datasets. Features such as the texture of raw material and the partner company can be extracted from a static dataset. From the event log, the trace $\sigma = \langle e_1, \ldots, e_n \rangle$ is extracted. Subsequently, three temporal attributes of elapsed time ($x_E$), lagged time ($x_L$), and the day of the week ($x_D$), are derived from the timestamp of each event to further reflect the sequential nature of traces. The elapsed and lagged time of $i$-th event ($e_i$) can be denoted by:

$$x_E(e_i) = \begin{cases} 0, & if \ i = 1 \\ \pi_t(e_i) - \pi_t(e_1), & otherwise \end{cases} \quad (5)$$

$$x_L(e_i) = \begin{cases} 0, & if \ i = 1 \\ \pi_t(e_i) - \pi_t(e_{i-1}), & otherwise \end{cases} \quad (6)$$

These two features enable the analysis of the relationships and patterns between observations at different times. Lastly, the day of the week of the $i$-th event $x_D(e_i)$ can be extracted from the timestamp as a categorical value. Three time-related features can be extracted by repeating these computations for all events in a procurement trace of a pipe spool.

### 3.2 Procurement Lead Time Prediction

In our proposed framework, mainly two types of datasets are utilized. The static dataset and event log are processed independently to predict PLT, and their representations are combined only in the final fully connected layer. First, the trace and time-related features are mapped into another vector using bidirectional long short-term memory (Bi-LSTM) cells, which capture contextual dependencies from both past and future events. Let $x_t = (e_t, x_E(e_t), x_L(e_t), x_D(e_t))$ denote an event and its time-related features at step $t$. The hidden representation of a trace is updated by:

$$c_{fwd,t}, h_{fwd,t} = LSTM_{fwd}(x_t) \quad (7)$$

$$c_{bwd,t}, h_{bwd,t} = LSTM_{bwd}(x_t) \quad (8)$$

where $LSTM_{fwd}$ is forward-pass and $LSTM_{bwd}$ is backward-pass LSTM layer respectively. Subsequently, the static dataset is linearly transformed into a dense vector. Let $x_s$ be the static features of a pipe spool. The representation $h_s$ of a pipe spool can be obtained by:

$$h_s = F_s(x_s; \boldsymbol{W}_s, \boldsymbol{B}_s) \quad (9)$$

where the function $F_s$ denotes multi-layered perceptron network with ReLU [11] activation function. $W_s$ and $B_s$ represent the weights and the corresponding biases for transformation respectively. Based on the representations $h_{fwd,t}, h_{bwd,t}$ and $h_s$, the final fully connected layers (FC) process concatenated representations to predict PLT, which can be denoted by:

$$\hat{y} = FC([h_s|h_{fwd,t}|h_{bwd,t}]; W_{fc}, B_{fc}) \tag{10}$$

where $W_{fc}$ and $B_{fc}$ are the set of trainable parameters and biases respectively.

## 4 Experiments

### 4.1 Experimental Settings

Table 1. Statistical description of dataset.

| Num. of pipe spools | Num. of static attrs. | Num. of dynamic attrs. | Num. of events | Trace length |
|---|---|---|---|---|
| 106,403 | 12 | 5 | 2,919,961 | 18-36 |

Table 1 presents the summary of statistics for the static dataset and the event log. In addition to PLT, we evaluated pipe production lead time and post-processing lead time against existing approaches [5, 6] to validate our method in more detail. Notably, the summation of production and post-processing lead time equals PLT. The baseline methods are implemented precisely as described in their papers and rely exclusively on static datasets. Performance is assessed in terms of mean absolute error (MAE), root mean squared error (RMSE) and mean absolute percentage error (MAPE). The computation costs are measured in terms of time. 70% of pipe spools are utilized for training, 10% for validation, and 20% for testing.

### 4.2 Experimental Results

**Performance Evaluation.** To evaluate the effectiveness of the consideration of process dynamics in PLT prediction, two performance evaluations are conducted. First, we compare the proposed methods with existing approaches that rely solely on static features, in terms of both predictive performance and computational cost. Second, we perform a performance comparison against representative sequential architectures. This includes recurrent neural networks (RNNs) [12], LSTM [10], Bi-LSTM [10], gated recurrent units (GRUs) [13], Bi-GRU [13], and Transformers [14].

Table 2 below demonstrates that the proposed method consistently outperforms all baselines for every lead time category and every performance metric. Our performances are reported with reference to the Bi-LSTM model. Regarding accuracy, we achieved an MAE of 4.18 days for production, 1.66 days for post-processing, and 4.66 days for procurement, representing an additional 22.6% to 50.4% reduction rela-tive to the best baseline DT. A similar trend holds for RMSE, indicating that both tails of the

**Table 2.** Predictive performance evaluation of the proposed method and baseline methods.

|  | Production Lead Time | | | Post Processing Lead Time | | | Procurement Lead Time | | |
|---|---|---|---|---|---|---|---|---|---|
|  | MAE | RMSE | Cost | MAE | RMSE | Cost | MAE | RMSE | Cost |
| LR | 7.37 | 9.61 | 15.38 | 4.71 | 6.38 | 15.82 | 8.41 | 10.69 | 15.33 |
| PLS | 7.24 | 9.48 | 12.94 | 4.68 | 6.39 | 20.23 | 8.30 | 10.56 | 12.50 |
| DT | <u>5.45</u> | 8.92 | 7.53 | <u>3.35</u> | 5.91 | 9.65 | <u>6.02</u> | 9.84 | 7.28 |
| RF | 9.55 | 11.95 | 0.39 | 6.32 | 8.26 | 0.43 | 10.54 | 12.98 | 0.41 |
| MLP | 6.29 | <u>8.45</u> | 186.63 | 4.02 | <u>5.64</u> | 463.22 | 7.37 | <u>9.62</u> | 198.59 |
| **Ours** | **4.18** | **6.44** | 458.07 | **1.66** | **3.21** | 1025.25 | **4.66** | **7.13** | 794.13 |

*MAE and RMSE are reported in days, MAPE expressed as a percentage (%)

error distributions are effectively suppressed.

Notably, all performance evaluation metrics achieve the best performance, highlighting the proposed method's reliability for industrial deployment where relative error is crucial. These gains stem from our process-aware framework, which captures the flow of process and temporal dynamics of the pipe spool procurement. However, the computational cost is relatively higher, primarily because our model processes higher-dimensional event log and employs more complex architecture. This represents a trade-off between accuracy and efficiency. Nevertheless, it is worth noting that the additional computational burden remains acceptable in practical settings, given the substantial improvements in predictive performance.

**Table 3.** Predictive performance comparison across different sequential architectures.

|  | Production Lead Time | | | Post Processing Lead Time | | | Procurement Lead Time | | |
|---|---|---|---|---|---|---|---|---|---|
|  | MAE | RMSE | Cost | MAE | RMSE | Cost | MAE | RMSE | Cost |
| RNN | 4.17 | 6.34 | 651.84 | 2.22 | 3.80 | 2054.20 | 4.66 | 7.03 | 1179.75 |
| LSTM | 4.20 | 6.38 | 844.09 | 1.71 | 3.23 | 1615.85 | 4.45 | **6.87** | 598.42 |
| GRU | 4.16 | 6.37 | 525.44 | 1.72 | 3.32 | 760.22 | 4.46 | 6.92 | 671.90 |
| Bi-LSTM | 4.18 | 6.44 | 719.61 | **1.66** | **3.21** | 1361.54 | **4.44** | 6.88 | 598.24 |
| Bi-GRU | **4.16** | **6.36** | 1074.54 | 1.76 | 3.34 | 1102.06 | 4.66 | 7.13 | 807.34 |
| Transformer | 4.26 | 6.46 | 1249.74 | 2.13 | 3.65 | 999.10 | 4.91 | 7.35 | 455.63 |

*MAE and RMSE are reported in days, MAPE expressed as a percentage (%)

Table 3 presents the predictive performance of different sequential architectures for production lead time, post-processing lead time, and procurement lead time. Overall, recurrent architectures such as RNN, LSTM, and GRU show competitive accuracy across all lead time categories, with GRU and Bi-LSTM yielding consistently lower

MAE and RMSE in post-processing and procurement lead time prediction. Bi-GRU achieves the best performance in production lead time prediction, recording the lowest MAE (4.16 days) and RMSE (6.36 days). Likewise, Bi-LSTM achieves the best balance between accuracy and efficiency, attaining the lowest MAE (1.66 days) and RMSE (3.21 days) for post-processing lead time, while also performing competitively in PLT.

**Ablation Study.** To further assess the contribution of different feature groups, ablations studies are conducted by selectively removing time-related features (TRF) and event log features (EL).

Table 4. Ablation study results.

|  | Production Lead Time | | | Post Processing Lead Time | | | Procurement Lead Time | | |
| --- | --- | --- | --- | --- | --- | --- | --- | --- | --- |
|  | MAE | RMSE | MAPE | MAE | RMSE | MAPE | MAE | RMSE | MAPE |
| Full | 4.18 | 6.44 | 0.19 | 1.66 | 3.21 | 0.12 | 4.44 | 6.88 | 0.11 |
| w/o TRF | 5.98 | 8.29 | 0.28 | 4.62 | 6.74 | 0.30 | 8.11 | 10.58 | 0.21 |
| w/o EL | 6.04 | 8.41 | 0.30 | 5.90 | 7.99 | 0.46 | 10.70 | 13.53 | 0.30 |

*MAE and RMSE are reported in days, MAPE expressed as a percentage (%)

Table 4 shows the removal of either feature group leads to substantial degradation in predictive performance across all lead time categories. When time-related features are excluded (w/o TRF), the prediction error increases notably, with MAE rising from 4.18 to 5.98 days in production lead time and from 1.66 to 4.62 days in post-processing lead time. A similar trend is observed for PLT, where MAE increases from 4.44 to 8.11 days. These results underscore the critical role of temporal information in capturing process dynamics and ensuring accurate prediction.

The exclusion of event log (w/o EL) results in an even more pronounced performance drop. In post-processing lead time prediction, the MAE nearly quadruples from 1.66 to 5.90 days, while in procurement lead time, it increases from 4.44 to 10.70 days. The MAPE values also increase sharply, indicating that process-specific contextual information in event logs is indispensable for precise lead time estimation.

### 4.3 Discussion

The experimental results show that a process-aware approach that explicitly encodes temporal dynamics and event log structures provides meaningful gains in both predictive accuracy and operational reliability for pipe spool PLT forecasting. Against existing methods, the proposed method achieves a 22.6%–50.4% reduction of MAEs relative to the best static baseline (DT). RMSE exhibits a similar pattern with a 27.5%–45.7% reduction, indicating not only lower average error but also suppression of tail errors. We attribute these improvements to the model's ability to capture process and temporal dependencies, reducing both bias and variance in forecasts. At the same time, we observe a computational trade-off. Since the adoption of event log exhibits higher dimension, the computational cost is higher than for purely static models. Nevertheless, from a deployment perspective, the substantial accuracy gains justify the added cost in periodic inference settings typical of industrial operations.

The comparison across sequential architectures (Table 3) provides two practical insights. 1) Bidirectionality: Bi-GRU attains the best production PLT accuracy (lowest MAE/RMSE), while Bi-LSTM achieves the best post-processing PLT performance and remains competitive for procurement. These results indicate that incorporating both past and future process contexts is relevant for workflows where sequential dependencies extend in both directions. 2) Recurrent Structure: Although the Transformer demonstrates strong representational capability, its performance is less stable across different lead time categories and is accompanied by a significantly higher computational cost. This indicates that, for industrial deployment where both predictive accuracy and efficiency are critical, lightweight recurrent architectures may offer more practical advantages.

The ablation study (Table 4) clarifies where the gains come from. Specifically, it highlights that both time-related features and the consideration of process dynamics are essential for robust performance. Their complementary roles, which are temporal continuity on one hand and process context on the other, form the foundation of the proposed framework's predictive accuracy. Thus, temporal continuity and process context are complementary, and removing either induces performance loss.

## 5   Conclusion

Rapid response to procurement delays is crucial in the engineered-to-order industry of shorter construction cycles. This study presents a predictive framework that enables early detection of procurement delays by forecasting procurement lead time (PLT), specifically focusing on pipe spools. Three time-related features and additional characteristics were extracted from an event log that records every activity and attribute in the procurement workflow. The architecture is designed to learn not only these features but also static spool features. The experimental result shows that the proposed method performs best on every evaluation metric for three prediction tasks: production, post-processing, and procurement lead time. The findings demonstrate that the decision window available to production planners directly mitigates schedule delays and cost overruns. Predicting and analyzing pipe workloads in advance balances the load across fabrication and post-processing vendors and prevents excessive concentration, alleviating bottlenecks. The proposed method also offers a scalable analytics pipeline with the same feature engineering and sequential learning procedure that can be utilized in other engineered-to-order domains where lead times drive project risk.

There are certain limitations that our proposed method is subject to. First, the domain generalization beyond the development remains to be validated under heterogeneous suppliers, contracts, and exogenous shocks such as weather and congestion. Furthermore, we evaluate statistical accuracy primarily through MAE/RMSE and do not examine how improvements in accuracy translate into operational cost savings (e.g., schedule quality, delay penalties, and expediting costs). Future work should explore: 1) multimodal fusion that considers perspectives such as supplier reliability, weather, and logistics tracking. 2) estimating the expected cost impact and combining the model with optimization techniques to derive optimal schedules.

**Acknowledgement.** This work was supported by the National Research Foundation of Korea (NRF) grant funded by the Korea government (MSIT). (No. RS-2023-00208999) (RS-2023-00242528).